\documentclass[conference]{IEEEtran}
\IEEEoverridecommandlockouts
\usepackage{cite}
\usepackage{amsmath,amssymb,amsfonts}
\usepackage{algorithmic}
\usepackage{graphicx}
\usepackage{multirow}
\usepackage{booktabs}
\usepackage{textcomp}
\usepackage{xcolor}
\usepackage{url}
\usepackage{verbatim}
\usepackage{todonotes}
\usepackage{fancyhdr}

\def\BibTeX{{\rm B\kern-.05em{\sc i\kern-.025em b}\kern-.08em
    T\kern-.1667em\lower.7ex\hbox{E}\kern-.125emX}}
    
\begin{document}

\title{Integrated Convolutional and Recurrent Neural Networks for Health Risk Prediction using Patient Journey Data with Many Missing Values \\
}

\author{
    \IEEEauthorblockN{
        Yuxi Liu\thanks{Yuxi Liu is the corresponding author.}\IEEEauthorrefmark{1},
        Shaowen Qin\IEEEauthorrefmark{1},
        Antonio Jimeno Yepes\IEEEauthorrefmark{2},
        Wei Shao\IEEEauthorrefmark{3},
        Zhenhao Zhang\IEEEauthorrefmark{4},
        Flora D. Salim\IEEEauthorrefmark{5},
    }
    \IEEEauthorblockA{
        \IEEEauthorrefmark{1}College of Science and Engineering, Flinders University, Adelaide, SA, Australia\\
         \IEEEauthorrefmark{2}School of Computing Technologies, RMIT University, Melbourne, Victoria, Australia\\
        \IEEEauthorrefmark{3}College of Electrical and Computer Engineering, UC Davis, Davis California, CA, USA\\
        \IEEEauthorrefmark{4}College of Life Sciences, Northwest A\&F University, Yangling, Shaanxi, China\\
        \IEEEauthorrefmark{5}School of Computer Science and Engineering, UNSW, Sydney, NSW, Australia \\
            \{liu1356, shaowen.qin\}@flinders.edu.au
            antonio.jose.jimeno.yepes@rmit.edu.au
            weishao@ucdavis.edu\\
            zhangzhenhow@nwafu.edu.cn
            flora.salim@unsw.edu.au
    }
}

\maketitle
\begin{abstract}
Predicting the health risks of patients using Electronic Health Records (EHR) has attracted considerable attention in recent years, especially with the development of deep learning techniques. Health risk refers to the probability of the occurrence of a specific health outcome for a specific patient. The predicted risks can be used to support decision-making by healthcare professionals. EHRs are structured patient journey data. Each patient journey contains a chronological set of clinical events, and within each clinical event, there is a set of clinical/medical activities. Due to variations of patient conditions and treatment needs, EHR patient journey data has an inherently high degree of missingness that contains important information affecting relationships among variables, including time. Existing deep learning-based models generate imputed values for missing values when learning the relationships. However, imputed data in EHR patient journey data may distort the clinical meaning of the original EHR patient journey data, resulting in classification bias. This paper proposes a novel end-to-end approach to modeling EHR patient journey data with Integrated Convolutional and Recurrent Neural Networks. Our model can capture both long- and short-term temporal patterns within each patient journey and effectively handle the high degree of missingness in EHR data without any imputation data generation. Extensive experimental results using the proposed model on two real-world datasets demonstrate robust performance as well as superior prediction accuracy compared to existing state-of-the-art imputation-based prediction methods.
\end{abstract}


\section{Introduction}
The availability of large amounts of electronic health records (EHR) has led to increased research interest in the application of artificial intelligence-based approaches to health risk prediction. Accurate risk prediction of a specific health outcome can be used to support decision-making by healthcare professionals. Examples of research studies in this area include disease risk prediction \cite{ye2020lsan}, mortality risk prediction \cite{liu2022modeling}, and patient subtyping classification \cite{yin2020identifying}. These studies usually construct deep neural networks (DNNs) that combine recurrent neural networks (RNNs) and attention, including time-aware mechanisms, to learn patient representations from EHR data for health risk predictions.

Patient representation learning refers to a dense mathematical representation of a patient characterized by representing valuable information from EHR data \cite{si2021deep}. The overall structure of EHR is built upon patient journeys. A patient journey is a record in which a set of time-ordered clinical events involving a single patient are chained together, and within each clinical event, there is a set of clinical/medical activities. Fig. 1 provides an overview of this structure using the MIMIC-III dataset \cite{johnson2016mimic} as an example. In this example, the patient's records have serious irregularity problems, e.g., large numbers of missing values. The high degree of missingness of EHR patient journey data dramatically increases the difficulty of directly applying existing state-of-the-art DNNs \cite{ozyurt2021attdmm, ibrahim2021knowledge, theis2021improving} to achieve desirable learning outcomes.
\begin{figure}[!htb]
        \centering
        \includegraphics[width = 1.0\linewidth]{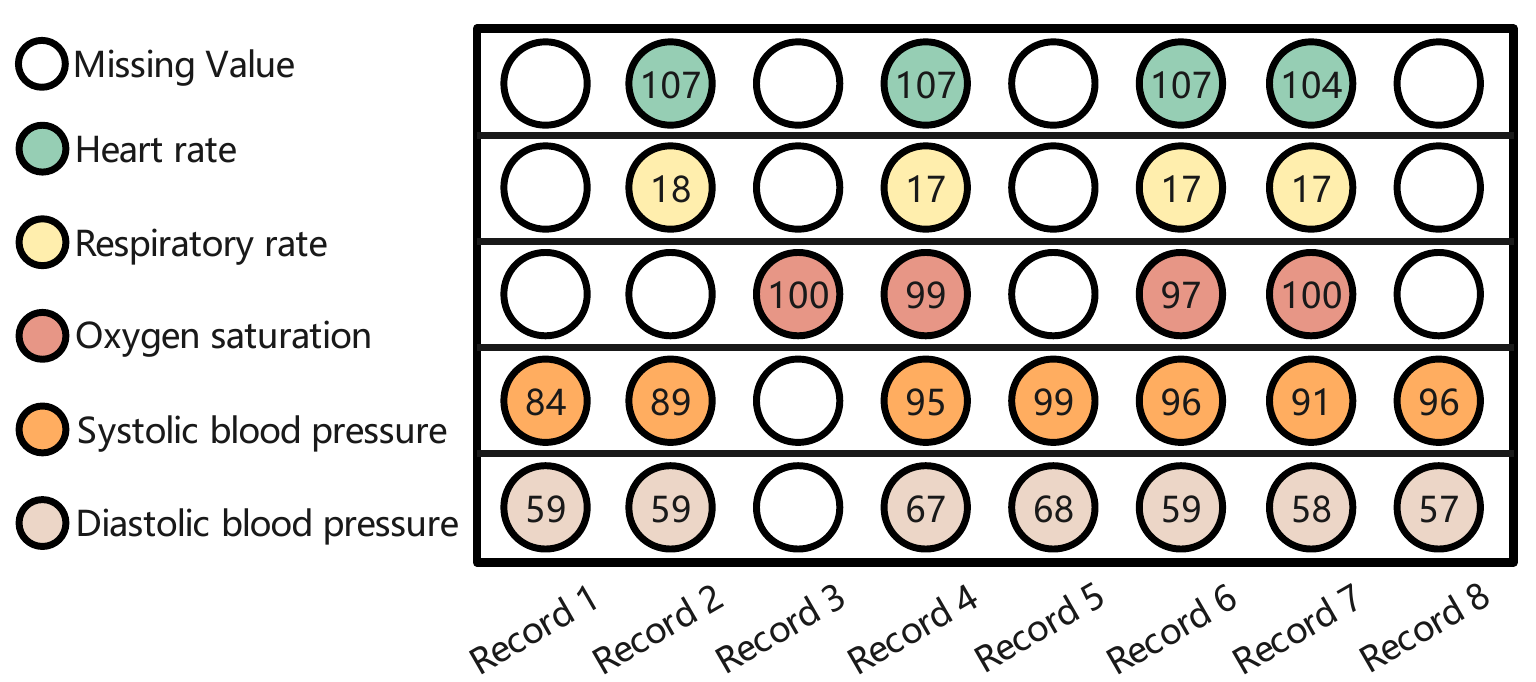}
        \caption{\textbf{An example of a patient's clinical records. The physician conducts/prescribes the necessary lab tests each time a patient is seen.}}
        \label{fig:DataFrame}
\end{figure}

Much of the research up to now has focused on utilizing imputation methods to address the missing values in EHRs \cite{jakobsen2017and, harutyunyan2019multitask, sheikhalishahi2020benchmarking, zhang2021missing}. These studies regard the problem of missing value imputation as data generation, and the generated imputation values are used to create a complete data matrix that can be analyzed using machine learning or statistical models.

Despite their efficacy, imputation methods suffer one significant drawback for healthcare-related applications, that is, imputed data in EHR patient journey data may distort the clinical meaning of the original EHR patient journey data. Due to changes in symptoms, physicians often selectively examine a patient's vital signs while continuously monitoring the patient \cite{tan2020data}. When a certain symptom disappears, corresponding indicators are no longer examined. This results in missing values. Therefore, the missing pattern of clinical events matters significantly in healthcare, where the different categories of missing data may reflect a typical clinical symptom that contains important information affecting the relationships at the variable level and the time dimension. Building a more adaptive method to learn the impact of missing values would greatly improve the targeted prediction.

Inspired by representation learning-related models \cite{wang2019inpatient2vec, si2021deep, liu2022modeling}, we propose an end-to-end, novel, and robust model by modeling EHR patient journey data with Integrated Convolutional and Recurrent Neural Networks. The intuitions behind Integrated Convolutional and Recurrent Neural Networks can be seen as modeling long- and short-term temporal patterns between clinical records within each patient's journey data. The long-term temporal patterns between clinical records model how each clinical record relates to the other records in the complete EHR patient journey. The short-term temporal patterns between clinical records model how every clinical record relates to each other in a short period of time. The modeling of these temporal patterns provides richer patient journey representations, promoting accurate modeling of patient journeys and leading to good prediction performance.

In the field of computer vision and short-term load forecasting (e.g., electricity consumption and traffic jam situation), convolution neural network (CNN) based prediction models have achieved superior performance by successfully extracting the local trend and the same pattern from input images and time series \cite{lai2018modeling, yamashita2018convolutional, tian2018deep}. Motivated by these successful applications, we design a one-dimensional CNN (1D-CNN) to effectively model short-term temporal patterns between clinical records within each patient's journey data. The 1D-CNN is designed with a simple and effective kernel size configuration that is able to elegantly handle EHR patient journeys with a high degree of missingness without imputation data generation. Then, we utilize Gated Recurrent Units (GRU) \cite{cho2014learning} to model long-term temporal patterns between clinical records within each patient's journey data.

We validate our proposed neural network on mortality prediction tasks from two publicly available EHR datasets that have a large degree of missing values. The results indicate that our model outperforms state-of-the-art imputation-prediction models by large margins.

\section{RELATED WORK}
Early examples of research into the imputation method include \cite{acuna2004treatment, kang2013prevention, jakobsen2017and}. These studies apply statistical methods to multivariate time series (MTS) data, such as the case deletion method as well as mean and median imputation methods, which handle missing values but largely ignore correlations variables. The study by \cite{honaker2010missing} offers a new model for imputing missing values that can take into consideration the variable correlations. The methodological approach taken in the study \cite{honaker2010missing} is built on a statistical model to extract smooth time trends, shifts across cross-sectional units, and correlations over time and space from MTS data. The extracted features are used to impute missing values. The imputed values are then used to construct complete data matrices fed into predictive models.

In another major study, 3D-MICE \cite{luo20183d} combines the MICE method and Gaussian Process to impute missing values. This kind of construction is particularly utilized to integrate temporal and cross-variable information and inspire the design principle of later works like T-LGBM \cite{xu2019estimating} and ELMV \cite{liu2020elmv}. T-LGBM extracts temporal and cross-variable features as inputs, then fed into the LightGBM method to impute missing values. ELMV derives multiple subsets from MTS data, which have a smaller degree of missingness than the overall dataset. The subsets are fed into the XGBoost method to implement an ensemble of classifiers that contributes to reducing the bias caused by a large number of missing values. The learning process integrates temporal and cross-variable information from subsets, leading to more robust model outputs.

More recent work has focused on deep learning-based imputation methods. Examples of representative deep imputation methods include BRNN \cite{suo2019recurrent}, CATSI \cite{yin2019context}, BRITS \cite{cao2018brits}, InterpNet \cite{shukla2019interpolation}, and GRU-D \cite{che2018recurrent}. These deep imputation methods have mainly been applied to MTS data imputation.
Given MTS data, BRNN generates the imputed values for each variable with the last observed value or the mean values of the same variable. These imputed values are used as initial imputed values for the complete data matrix, fed into a bidirectional RNN to predict the final values. CATSI comprises a context-aware recurrent imputation and a cross-variable imputation, which are used to capture temporal information and cross-variable relations from MTS data. A fusion layer in CATSI is used to integrate these two imputation outputs into the final imputation outputs. BRITS employs a bidirectional RNN to impute missing values in MTS data and then exploits these imputed values to predict the final imputed values. The two prediction losses are tuned together in BRITS. InterpNet comprises an interpolation network and a prediction network. The interpolation network is an unsupervised learning network that can be used to impute missing values of MTS data. The prediction network is used to generate prediction results. The overall structure of GRU-D is built upon GRU. GRU-D mainly incorporates the empirical mean value and the previous observation to impute missing values.

It is worth noting that InterpNet and GRU-D take the irregular interval of MTS data into consideration when imputing missing values. InterpNet mainly converts observed records into equally spaced. Despite its efficacy, the conversion process inevitably leads to information loss due to variable-length observations. Due to heterogeneity among patients (e.g., age, clinical history), the corresponding number of observed records varies widely. GRU-D introduces observed records and corresponding timestamps into GRU to impute missing values as the decay of previous input values toward the overall mean/sampling over time. The time-decay mechanism used in GRU-D continues to be used by BRITS \cite{cao2018brits} and CATSI \cite{yin2019context}.

Compared with all the aforementioned imputation models, our proposed model has the following advantages: (1) It is a general model that can be used to create useful predictions for a variety of health risk assessment scenarios. (2) It takes into account long- and short-term temporal patterns between clinical records within each patient's journey data. (3) It handles missing data in each patient's journey data without imputation data generation.


\section{METHODS}
\subsection{Basic Notations}
EHR data consists of patients' time-ordered records. Each patient's records ensemble can be further categorized as a patient journey. The EHR patient journey data is denoted by $X^{p}$ = $[x_{1}^{p}, \cdots, x_{t}^{p}, \cdots, x_{T_{p}}^{p}]$ $\in$ $\mathbb{R}^{N \times T_{p}}$, where $N$ is the number of sequential dynamic features (that occur over time, e.g., vital signs) and $T_{p}$ is the number of records. For simplicity, we drop the $p$ when it is unambiguous in the following sections.

\subsection{Model Architecture}
The architecture of the proposed model is shown in Fig. 2. The proposed model comprises a 1D-CNN and a GRU recurrent component. 
\begin{figure*}[!htb]
        \centering
        \includegraphics[width = 1.0\linewidth]{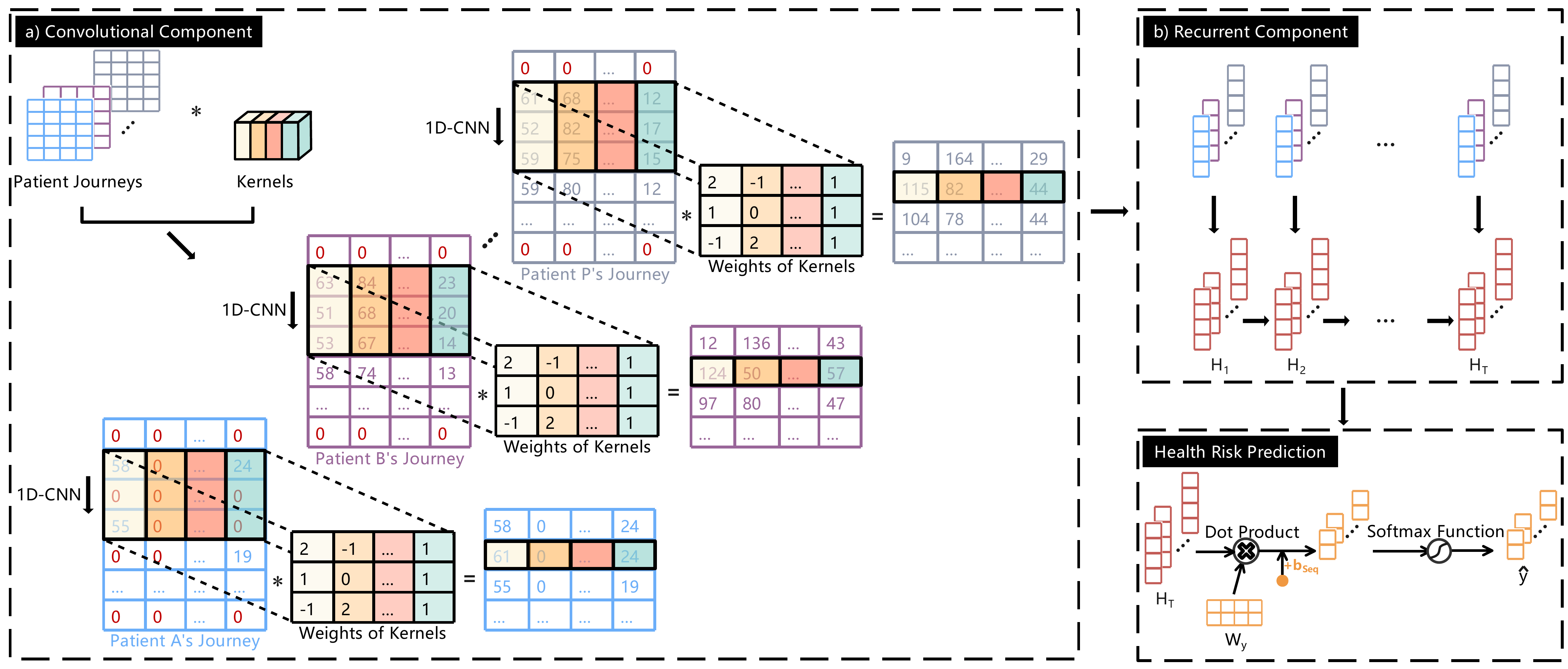}
        \caption{\textbf{Architecture of the proposed model.}}
        \label{fig:overview}
\end{figure*}

\subsubsection{Convolutional Component}
Before implementing the proposed convolutional component, we introduce the background information on convolution operation. The convolution operation can be seen as an element-wise multiplication of a filter (kernel) with its corresponding local window (receptive field) in the input feature map, as shown in Fig. 2a. 

The convolutional component is a 1D-CNN that has a simple and effective kernel size configuration. We build a convolution kernel for each sequential dynamic feature. Each kernel has a size of 3 and a stride of 1. All patient journey modeling shares a set of convolutional kernels.
For example, given an input [58, 0, 55] (zero is the prefilled value) and a set of weights of a kernel [2, 1, -1] (see Fig. 2a), the dense record representation in patient A's journey is calculated as [61, $\cdots$], where 61 = 58 $\cdot$ 2 + 0 $\cdot$ 1 + 55 $\cdot$ -1. The number of weights for a kernel is determined by the size of the kernel (e.g., set to 3).
In the following, we provide the implementations details.

The 1D-CNN models the causative and associative relationships between consecutive records in each patient's journey data. The benefit of using a kernel of size 3 is that it can integrate each record and its adjacent records (records immediately before and after each record) into an overall record representation, termed a new dense record representation. For example, if one patient journey data contains 60 records, 60 dense record representations would be produced after training. 
The kernel size is a hyperparameter that might need to be adjusted depending on the dataset.

When using the 1D-CNN, we pad the data by embedding a zero vector before the first record of $X$ and after the last record of $X$. The use of zero vectors is to keep the shape of patient journey data after the convolution operation consistent with its shape before the convolution operation. $X = [x_{1}, x_{2}, \cdots, x_{T}]$ is now expanded into $X^{\prime} = [x_{0}, x_{1}, \cdots, x_{t}, \cdots, x_{T}, x_{T+1}]$. $x_{0}, x_{T+1} \in \mathbb{R}^{N}$ are zero vectors. Then, we apply 1D convolution operation only over the horizontal dimension of $X^{\prime}$ (see Fig. 1, X-axis). We use a combination of $N$ kernels $\{W_{n}\}^{N}_{n = 1}$ to handle $N$ sequential dynamic features. For example, $X^{\prime}_{t - 1: t + 1}$ represents the concatenation of records $x_{t-2}$, $x_{t-1}$, and $x_{t}$ from $X^{\prime}_{t - 1}$ to $X^{\prime}_{t + 1}$. A kernel $W^{(n)} \in \mathbb{R}^{3 \times 1}$ is naturally applied on the window of $X^{\prime (n)}_{t - 1: t + 1}$ to produce a new sequential dynamic feature $z_{t}^{(n)} \in \mathbb{R}$ with a rectified linear unit (ReLU)~\cite{nair2010rectified} activation function as:
\begin{equation}
\begin{split}
\label{eq:1}
z^{(n)}_{t} = ReLU (X^{\prime (n)}_{t -1: t+ 1} \cdot W^{(n)} + b^{(n)}),
\end{split}
\end{equation}
where $b^{(n)} \in \mathbb{R}$ is a bias term and $ReLU(k) = max(k, 0)$. This kernel is applied to each possible window of values in the whole description $X_{1:3}^{\prime(n)}$, $X_{3:5}^{\prime (n)}$, $\cdots$, $X_{T:T + 2}^{\prime (n)}$ to generate a feature map $z^{(n)} \in \mathbb{R}^{T}$ as: $z^{(n)} = [z_{1}^{(n)}, z_{2}^{(n)}, \cdots, z_{T}^{(n)}]$, n = 1, 2, $\cdots$, N. Each kernel produces a sequential dynamic feature. Since we have $N$ kernels, the final vector representation of a patient journey can be obtained by concatenating all these features, $z \in \mathbb{R}^{N \times T}$.

\subsubsection{Recurrent Component}
Through the aforementioned 1D-CNN implementation, the proposed model would model short-term temporal patterns within each patient's journey data and it would address missing data without imputation data generation. The outputs of 1D-CNN are fed into the recurrent component (GRU \cite{cho2014learning}) to model their long-term temporal patterns (see Fig. 2b). Given input $z \in \mathbb{R}^{N \times T}$ and previous hidden state $H_{t - 1} \in \mathbb{R}^{g}$, the current hidden state $H_{t}$ can be obtained through the following steps.

We feed $z_{t}$ and $H_{t-1}$ into a gating mechanism. The gating mechanism, including a reset gate $R_{t}$ and an update gate $U_{t}$, is trained to decide which of the previous information will be retained for $H_{t}$. The functions of the gating mechanism are:
\begin{equation}
\begin{split}
\label{eq:2}
R_{t} = \sigma (W_{R} \cdot [H_{t - 1}, z_{t}] + b_{R}), \\
U_{t} = \sigma (W_{U} \cdot [H_{t - 1}, z_{t}] + b_{U}),
\end{split}
\end{equation}
where $W_{R} \in \mathbb{R}^{g \times (N + g)}$ and $W_{U} \in \mathbb{R}^{g \times (N + g)}$ are learnable parameters. $b_{R} \in \mathbb{R}^{g}$ and $b_{U} \in \mathbb{R}^{g}$ are biases. $\sigma$ is the sigmoid activation function that is used to normalize the outputs $R_{t}$ and $U_{t}$ in $[0, 1]$. The $z_{t}$ and the element-wise multiplication of $H_{t-1}$ with $R_{t}$ are used to generate an intermediate $\widetilde{H}_{t}$. $H_{t}$ is obtained by the element-wise convex combinations between $\widetilde{H}_{t}$ and $U_{t}$ and has the following formula:
\begin{equation}
\begin{split}
\label{eq:3}
\widetilde{H}_{t} = tanh (W_{H} \cdot [R_{t} \odot H_{t - 1}, z_{t}] + h_{H}), \\
H_{t} = U_{t} \odot H_{t - 1} + (1 - U_{t}) \odot \widetilde{H}_{t},
\end{split}
\end{equation}
where $W_{H} \in \mathbb{R}^{g \times (N + g)}$ is a learnable parameter and $h_{H} \in \mathbb{R}^{g}$ is a bias. $\odot$ denotes the element-wise multiplication.

\subsubsection{Health Risk Prediction}
The generated last latent variable $H_{T}$ is fed into a softmax function to obtain the predicted $\hat{y}$. Its formula is:
\begin{equation}
\begin{split}
\label{eq:4}
\hat{y} = softmax(W_{y} \cdot H_{T} + b_{y}),
\end{split}
\end{equation}
where $W_{y}$ and $b_{y}$ are learnable parameters. The cross-entropy between the ground truth $y$ and the predicted $\hat{y}$ is used to calculate the loss. Accordingly, the objective function $\mathcal{L}$ of the health risk prediction task is the average of cross-entropy:
\begin{equation}
\begin{split}
\label{eq:5}
\mathcal{L} = - \frac{1}{P} \sum^{P}_{p = 1} (y_{p}^{\top} \cdot log (\hat{y}_{p}) + (1 - y_{p})^{\top} \cdot log (1 - \hat{y}_{p})),
\end{split}
\end{equation}
where $P$ is the number of patient journeys. $y_{p}$ is the ground-truth class/label for patient p's journey.

\section{EXPERIMENTS}
\subsection{Experimental Settings}
\subsubsection{Datasets and Tasks}
We validate the proposed model on the mortality risk prediction task from the MIMIC-III \cite{johnson2016mimic} and eICU dataset \cite{pollard2018eicu}. We use clinical times series (e.g., heart rate, glucose) as input for both the MIMIC-III~\cite{harutyunyan2019multitask} and eICU~\cite{sheikhalishahi2020benchmarking} datasets. 

In our work, we consider six binary classification tasks, three tasks per data set: 
\begin{itemize}
\item \textbf{In-hospital mortality (24 hours after ICU admission)} to evaluate ICU mortality based on the data from the first 24 hours after ICU admission.

\item \textbf{In-hospital mortality (36 hours after ICU admission)} to evaluate ICU mortality based on the data from the first 36 hours after ICU admission. 

\item \textbf{In-hospital mortality (48 hours after ICU admission)} to evaluate ICU mortality based on the data from the first 48 hours after ICU admission.
\end{itemize}

\subsubsection{Baselines}
We compare the proposed model with Mean, KNN, 3D-MICE, and Simple:
\begin{itemize}

\item \textbf{Mean} values of variables are used to impute the missing values.

\item \textbf{KNN} is the average values of the top $K$ most similar collections are used to impute the missing values.

\item \textbf{3D-MICE} \cite{luo20183d} is described in the related work section.

\item \textbf{Simple} concatenates the measurement with masking and time intervals, which are then fed into a predictor to make risk predictions \cite{che2018recurrent}.
\end{itemize}  

We compare the proposed model with BRNN \cite{suo2019recurrent}, CATSI \cite{yin2019context}, BRITS \cite{cao2018brits}, InterpNet \cite{shukla2019interpolation}, and GRU-D \cite{che2018recurrent} and GRU-D$_{t-}$ (without time decay mechanism). The outputs of Mean, KNN, 3D-MICE, Simple, BRNN, and CATSI are fed into GRU \cite{cho2014learning} to make final predictions. We also present one variant of our model (Ours$_{r-}$) that does not perform the recurrent component.

\subsubsection{Implementation Details \& Evaluation Metrics}
All the models are implemented using Python v3.9.7. We employ the following libraries: fancyimpute for KNN and MICE and PyTorch for the rest of the methods. The two datasets were randomly divided into training, validation, and testing sets for each task in a 70:15:15 ratio. The validation set is used to select the hyperparameters values. We repeat the training of all the methods ten times and report the average performance. Training and evaluations were performed on an NVIDIA A40 GPU with 48GB of memory. We augment the CrossEntropyLoss function with a class weight for highly imbalanced datasets. The area under the receiver operating characteristic curve (AUROC) and the area under the precision-recall curve (AUPRC) are used to evaluate classification models.

\subsection{Comparison with baselines}
\begin{table*}[htbp] \scriptsize
  \centering
  \caption{Comparison of performance on in-hospital mortality prediction using AUROC and AUPRC with variations of time window of an ICU stay between 24 to 48 hours.}
    \begin{tabular}{ccccccc}
    \toprule
    MIMIC-III/In-hospital Mortality Prediction & \multicolumn{2}{c}{24 hours after ICU admission} & \multicolumn{2}{c}{36 hours after ICU admission} & \multicolumn{2}{c}{48 hours after ICU admission} \\
    \midrule
    Metrics & AUROC & AUPRC & AUROC & AUPRC & AUROC & AUPRC \\
    \midrule
    Mean & 0.6780(0.017) & 0.2283(0.017) & 0.6821(0.016) & 0.2322(0.015)	& 0.6816(0.016) & 0.2314(0.015) \\
    KNN & 0.7122(0.016) & 0.2498(0.022) & 0.7057(0.015) & 0.2423(0.019) & 0.7086(0.013) & 0.2464(0.019) \\
    3D-MICE & 0.7016(0.019) & 0.2384(0.011) & 0.7198(0.020) & 0.2535(0.012)	& 0.7141(0.020) & 0.2319(0.013)  \\
    Simple & 0.6821(0.012) & 0.2315(0.010) & 0.6806(0.012) & 0.2307(0.010)	& 0.6791(0.013) & 0.2279(0.012) \\
    BRNN  & 0.6735(0.011) & 0.2037(0.012) & 0.6704(0.010) & 0.2023(0.012)	& 0.6732(0.011) & 0.2051(0.014) \\
    CATSI & 0.7042(0.011) & 0.2373(0.012) & 0.7024(0.013) & 0.2343(0.015)	& 0.7057(0.012) & 0.2379(0.012) \\
    BRIST & 0.7463(0.010) &  0.2880(0.016) & 0.7445(0.009) & 0.2856(0.016)	& 0.7447(0.009) & 0.2879(0.016) \\
    InterpNet & 0.6576(0.015) & 0.2113(0.024) & 0.6581(0.010) & 0.2198(0.019)	& 0.6559(0.008) & 0.2144(0.021) \\
    GRU-D & 0.7323(0.012) & 0.2821(0.014) & 0.7235(0.012) & 0.2679(0.015) & 0.7285(0.011) & 0.2763(0.015) \\
    GRU-D$_{t-}$ & 0.7137(0.011) & 0.2689(0.016) & 0.7342(0.015) & 0.2624(0.015) & 0.7244(0.011) & 0.2673(0.014) \\
    Ours & \textbf{0.8052(0.006)} & \textbf{0.3827(0.010)} & \textbf{0.8048(0.006)} & \textbf{0.3819(0.010)} & \textbf{0.8045(0.005)} & \textbf{0.3833(0.011)} \\
    Ours$_{r-}$ & 0.7438(0.003) & 0.3156(0.004) & 0.7436(0.002) & 0.3148(0.004)	& 0.7437(0.002) & 0.3157(0.004) \\
    \bottomrule
    \end{tabular}%
  \label{tab:addlabel}%
\end{table*}%
\begin{table*}[htbp] \scriptsize
  \centering
  \caption{Comparison of performance on in-hospital mortality prediction using AUROC and AUPRC with variations of time window of an ICU stay between 24 to 48 hours.}
    \begin{tabular}{ccccccc}
    \toprule
    eICU/In-hospital Mortality Prediction & \multicolumn{2}{c}{24 hours after ICU admission} & \multicolumn{2}{c}{36 hours after ICU admission} & \multicolumn{2}{c}{48 hours after ICU admission} \\
    \midrule
    Metrics & AUROC & AUPRC & AUROC & AUPRC & AUROC & AUPRC \\
    \midrule
    Mean & 0.7001(0.009) & 0.2711(0.011) & 0.7214(0.012) & 0.2720(0.015) & 0.7231(0.014) & 0.2628(0.012) \\
    KNN & 0.6899(0.009) & 0.2624(0.013) & 0.7117(0.011) & 0.2557(0.013)	& 0.7101(0.019) & 0.2467(0.017) \\
    3D-MICE & 0.6802(0.007) & 0.2507(0.008) & 0.7024(0.007) & 0.2399(0.009)	& 0.6895(0.008) & 0.2216(0.009)  \\
    Simple & 0.6946(0.015) & 0.2666(0.015) &	0.7173(0.012) & 0.2681(0.012) &	0.7195(0.012) & 0.2624(0.010) \\
    BRNN  & 0.6682(0.008) & 0.2395(0.009) & 0.6876(0.020) & 0.2407(0.025)	& 0.6783(0.012) & 0.2278(0.012) \\
    CATSI & 0.6956(0.013) & 0.2694(0.017) & 0.7119(0.017) & 0.2648(0.019)	& 0.7160(0.014) & 0.2617(0.013) \\
    BRIST & 0.7107(0.009) & 0.2570(0.008) & 0.7245(0.006) & 0.2577(0.010)	& 0.7254(0.006) & 0.2573(0.006) \\
    InterpNet & 0.7269(0.004) & 0.3308(0.010) & 0.7465(0.003) & 0.3116(0.017)	& 0.7537(0.003) & 0.2943(0.014) \\
    GRU-D & 0.6989(0.006) & 0.2659(0.007) & 0.7191(0.008) & 0.2689(0.008)	& 0.7190(0.012) & 0.2581(0.010) \\
    GRU-D$_{t-}$ & 0.6964(0.010) & 0.2726(0.013) & 0.7196(0.010) & 0.2628(0.013)	& 0.7202(0.015) & 0.2537(0.011) \\
    Ours & \textbf{0.7554(0.003)} & \textbf{0.3412(0.009)} & \textbf{0.7723(0.006)} & \textbf{0.3311(0.009)} & \textbf{0.7882(0.003)} & \textbf{0.3447(0.010)} \\
    Ours$_{r-}$ & 0.7257(0.004) & 0.2725(0.004) & 0.7394(0.002) & 0.2774(0.003)	& 0.7290(0.004) & 0.2547(0.003) \\
    \bottomrule
    \end{tabular}%
  \label{tab:addlabel}%
\end{table*}%
Tables III and IV show the performance of all approaches on two datasets. We find that the proposed model outperforms all baseline models on both evaluation metrics. For example, for the mortality prediction of MIMIC-III (48 hours after ICU admission), our proposed model achieves the highest AUROC with 0.8045 and standard deviation of 0.005. Similarly, the proposed model achieved the highest AUPRC with 0.3833 with a standard deviation of 0.011. 

As Tables III and IV show, deep imputation-prediction networks (e.g., BRIST, InterpNet, GRU-D) underperform other methods on these tasks. No significant differences in prediction performance are found between GRU-D and GRU-D$_{t-}$. 

Moreover, the proposed model (Ours) outperforms Ours$_{r-}$. It indicates that modeling of long-term temporal patterns between clinical records is a key step for superior prediction performance.

\section{DISCUSSION}
Prediction performance of our model over a window of 24 hours may offer some hints on possible research opportunities for early mortality prediction. Early prediction of patients at high risk of death might allow ICU staff to take appropriate mitigating actions to minimize risks and manage demand \cite{ozyurt2021attdmm}.

Perhaps the most interesting finding is the lower performance of imputation-prediction methods. This finding could have been generated by imputation bias since the imputed values might be biased due to the diversity of patient data, hence lacking reliability. 

The performance of the time decay mechanism is interesting but not surprising. This result corroborates the ideas of \cite{luo2020hitanet}, which suggested that the importance of features associated with each clinical event should not decay in a monotonic way. In other words, utilizing a time decay mechanism to deal with irregular time intervals of clinical events in longitudinal patient records might not be the most efficient/reasonable way.

\section{CONCLUSION}
In this paper, we have presented a novel deep learning method of integrating convolutional neural networks and recurrent neural networks for carrying out health risk prediction tasks using EHRs with a large number of missing values. Extensive experimental results demonstrate that our method is able to not only handle missing data in EHRs effectively but also outperform state-of-the-art imputation-prediction methods significantly. The present study will serve as a base for future studies, and the research method will be of broad use to the scientific and biomedical communities.


\end{document}